\documentclass[11pt]{article}

\usepackage[preprint]{acl}

\usepackage{times}
\usepackage{latexsym}
\usepackage{amsmath}   
\usepackage{amssymb}   
\usepackage{amsthm}
\usepackage{booktabs}
\usepackage{bm}

\newtheorem{definition}{Definition}

\usepackage[T1]{fontenc}

\usepackage[utf8]{inputenc}

\usepackage{enumitem}
\usepackage{booktabs}
\usepackage{array}
\usepackage{hyperref}
\usepackage{url}
\usepackage{multirow}
\usepackage{colortbl}
\usepackage{booktabs}
\usepackage{amssymb}
\usepackage{amsmath}
\usepackage{amsthm}
\usepackage{graphicx}
\usepackage{makecell}
\usepackage{threeparttable}
\usepackage{CJKutf8}
\usepackage{lscape}
\usepackage{fancyhdr}
\usepackage[ruled,vlined]{algorithm2e}

\usepackage{threeparttable}

\usepackage{microtype}

\usepackage{inconsolata}

\usepackage{graphicx}

\title{Orthogonal Low-rank Adaptation in Lie Groups for Continual Learning of Large Language Models}

\author{Kefan Cao \and Shuaicheng Wu \\
         Yingcai Honor College,\\ University of Electronic Science and Technology of China\\ 
\texttt{\normalsize kefancao@std.uestc.edu.cn}
}

\begin{document}
\maketitle
\begin{abstract}
Large language models (LLMs) suffer from catastrophic forgetting in sequential multi-task learning. Existing parameter regularization methods (e.g., O-LoRA, N-LoRA) mitigate interference via low-rank subspace orthogonality, but additive updates distort the intrinsic geometry of model parameters. We propose \textbf{OLieRA}, a Lie group–based fine-tuning framework that preserves parameter geometry through multiplicative updates while enforcing orthogonality across task subspaces. OLieRA achieves state-of-the-art performance on the Standard CL benchmark and remains highly competitive under large task sequences. It further inherits the replay-free and task-ID–free inference properties of O-LoRA, establishing a principled paradigm for continual learning in LLMs.
\end{abstract}

\section{Introduction}
Enabling lifelong continual learning in large language models (LLMs) is essential for real-world deployment, yet they remain highly susceptible to \textit{catastrophic forgetting}, whereby acquiring new tasks degrades performance on previously learned ones \citep{MCCLOSKEY1989109}. This limitation is especially severe in practical settings, where task adaptation (e.g., sentiment analysis) can impair prior capabilities such as technical document summarization \cite{kotha2024understanding}. \textit{Continual learning} addresses this challenge by enabling incremental knowledge acquisition from nonstationary task streams while preserving performance on historical tasks \cite{parisi2019continual}.

\begin{figure}[t]
    \small
    \centering
    \includegraphics[width=3.0in]{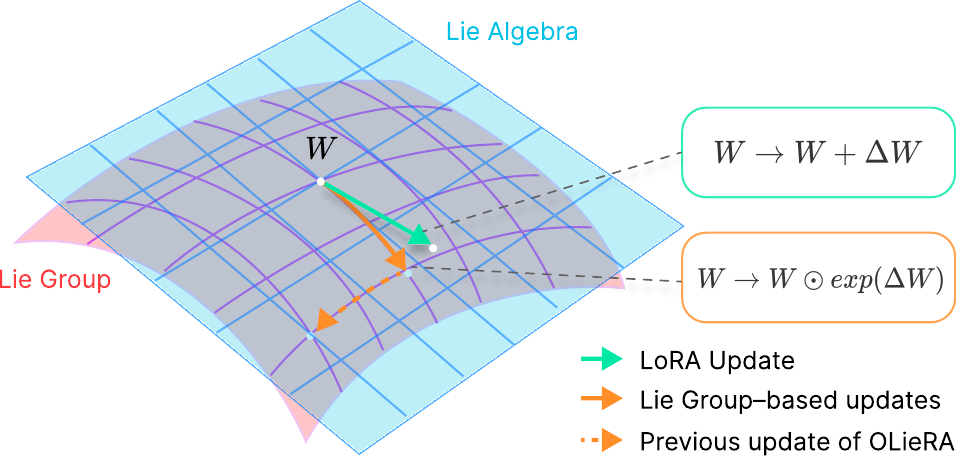}
    \caption{Illustration of the update mechanisms of OLieRA. While LoRA updates parameters, it overlooks the intrinsic parameter structure. In contrast, OLieRA incorporates a Lie group constraint in addition to orthogonality, thereby preserving the original parameter structure.}
    \label{fig:Update Methods}
\end{figure}

Existing continual learning methods fall into three paradigms, each with inherent constraints. \textbf{Replay-based} methods \cite{DBLP:journals/corr/abs-1906-01076,DBLP:journals/corr/Lopez-PazR17} rely on storing past data, incurring privacy risks and prohibitive computational costs for large LLMs. \textbf{Regularization-based} methods \cite{kirkpatrick2017overcoming} constrain parameter drift but degrade over long task sequences and struggle with ambiguous task boundaries \cite{peng2024scalable, ke2023continuallearningnaturallanguage}. \textbf{Architecture-based} methods \cite{razdaibiedina2023progressive, wang-etal-2023-rehearsal} expand or isolate parameters to reduce interference, but fragment model capacity and limit transfer to unseen tasks.

Our work builds on the regularization paradigm, extending Orthogonal Low-Rank Adaptation (O-LoRA) \cite{wang-etal-2023-orthogonal}. While O-LoRA mitigates interference via orthogonal gradient updates, its additive updates ($W \rightarrow W + \Delta W$) can distort the intrinsic geometry of LLM parameters \cite{si-etal-2025-generalized}. We propose \textbf{Orthogonal Low-Rank Adaptation under Lie Groups (OLieRA)}, which enforces orthogonality across full task subspaces and preserves geometric structure through Lie group–based updates \cite{si-etal-2025-generalized}, with minimal additional overhead and improved theoretical interpretability (Fig.~\ref{fig:Update Methods}).

OLieRA retains the key advantages of O-LoRA: \emph{privacy-friendliness} (no data replay), \emph{parameter efficiency}, and \emph{task-agnostic generalization}. It consistently outperforms prior orthogonal low-rank methods, including N-LoRA \cite{yang-etal-2025-parameter}.

The core contributions of this study are as follows:
\begin{enumerate}
    \item We introduce OLieRA, a Lie group–based regularization framework that achieves state-of-the-art results on standard continual learning benchmarks, surpassing O-LoRA and N-LoRA;
    \item We provide a geometric view of LLM fine-tuning, demonstrating that Lie group–constrained updates better preserve parameter structure, supported by extensive analytical experiments.
\end{enumerate}

\section{Related Work}
\subsection{Overview of Continual Learning}
Continual Learning \cite{ke2023continuallearningnaturallanguage, wang2023tracecomprehensivebenchmarkcontinual} aims to develop learning algorithms that accumulate knowledge from non-stationary data streams. In supervised continual learning, a sequence of tasks $\{\mathcal{D}_1, \dots, \mathcal{D}_T\}$ arrives sequentially. Each task $\mathcal{D}_t = \left\{ (x_i^t, y_i^t) \right\}_{i=1}^{n_t}$ contains an independent target dataset, where $x_i^t \in \mathcal{X}_t$ and $y_i^t \in \mathcal{Y}_t$. A single model must adapt to these tasks incrementally, and at the $t$-th task, it can only access $\mathcal{D}_t$. Typically, given a predictive model $h_\Theta$ parameterized by $\Theta$, the objective of continual learning is to optimize the following across all tasks:
\begin{equation}
\max_{\Theta} \sum_{k=1}^{T} \sum_{(x,y) \in \mathcal{D}_k} \log p_\Theta(y \mid x)
\end{equation}
where $p_{\Theta}(y|x)$ denotes the probability of predicting output $y$ given input $x$ under model $h_{\Theta}$.

\subsection{LoRA}
When adapting pre-trained models (PTMs) to specific tasks, it has been shown \cite{hu2021loralowrankadaptationlarge} that weight updates exhibit a low "intrinsic dimension." For a pre-trained weight matrix \( W \in \mathbb{R}^{d \times k} \), LoRA constrains updates via low-rank decomposition: \( W + \Delta W = W + BA \), where \( B \in \mathbb{R}^{d \times r} \), \( A \in \mathbb{R}^{r \times k} \), and the rank satisfies \( r \ll \min(d, k) \). During training, \( W \) remains fixed and does not receive gradient updates, while \( A \) and \( B \) contain trainable parameters. The modified forward pass for the computation \( h = Wx \) under LoRA becomes:
\[
h = Wx + \Delta W x = Wx + BAx
\]

\subsection{O-LoRA}
O-LoRA \cite{wang-etal-2023-orthogonal} is a continual learning framework for language models that combines instruction tuning with task-specific low-rank adaptation. For each new task, it incrementally adds a LoRA module while enforcing orthogonality between the parameter subspaces of the current and previous tasks to reduce interference.

For task $t$, O-LoRA learns LoRA parameters $\{A_t, B_t\}$ with $B_t \in \mathbb{R}^{d \times r_t}$. The task-specific update subspace is defined as
\begin{equation}
\mathcal{U}_t = \mathrm{span}\{\mathbf{b}_t^1, \ldots, \mathbf{b}_t^{r_t}\},
\end{equation}
where $\mathbf{b}_t^i$ denotes the $i$-th column of $B_t$. To prevent catastrophic forgetting, O-LoRA enforces orthogonality between subspaces by constraining
\begin{equation}
B_i^\top B_t = \mathbf{0}, \quad \forall i < t.
\end{equation}
This constraint is implemented via an orthogonality regularizer
\begin{equation}
L_{\text{orth}}(B_i, B_t) = \| B_i^\top B_t \|_F^2,
\end{equation}
which minimizes inter-task interference and preserves previously learned knowledge.

\subsection{N-LoRA}

In contrast to O-LoRA, N-LoRA \cite{yang-etal-2025-parameter} imposes a stronger constraint known as the non-parametric conflict condition.

\begin{definition}[Non-parametric Conflict]
    For all $(a,b)$, $\Delta W_1[a,b] \cdot \Delta W_2[a,b] = 0$,  
    where $\Delta W[a,b]$ denotes the element of $\Delta W$ at row $a$ and column $b$.  
    If matrices $\Delta W_1$ and $\Delta W_2$ satisfy this condition, they are said to be non-parametric conflicting at position $(a,b)$.
\end{definition}

Under this constraint, the previously updated parameter subspaces become more resistant to forgetting and sparser compared to O-LoRA, implying reduced parameter footprint and enabling more sequential adaptations. To reduce parameter collision, the authors apply $\ell_1$ regularization to the LoRA parameters $\Delta W_i$ of new tasks to enhance sparsity:
\begin{equation}
L_{\text{sparse}} = \lambda \|\Delta W_i\|_1
\end{equation}

\subsection{LieRA}

LieRA is a recently proposed parameter-efficient fine-tuning method. Unlike standard LoRA, which applies additive updates, LieRA \cite{si-etal-2025-generalized} models parameters as elements of a Lie group and performs updates in the associated Lie algebra, mapping them back via the exponential map. This yields a multiplicative update,
\begin{equation}
W \odot \exp(\Delta W),
\end{equation}
that explicitly preserves parameter structure. LieRA has shown consistent gains across vision and language tasks, including image understanding and commonsense reasoning. By maintaining the intrinsic structure of high-dimensional parameters (e.g., spatial locality in convolutional kernels), the Lie group–Lie algebra framework enables smooth, consistent updates, improves adaptability to new tasks, and retains the efficiency of low-rank adaptation while offering a principled geometric view of parameter updates.

\section{Theoretical Framework}

\begin{figure*}[t]
    \centering
    \includegraphics[width=6in]{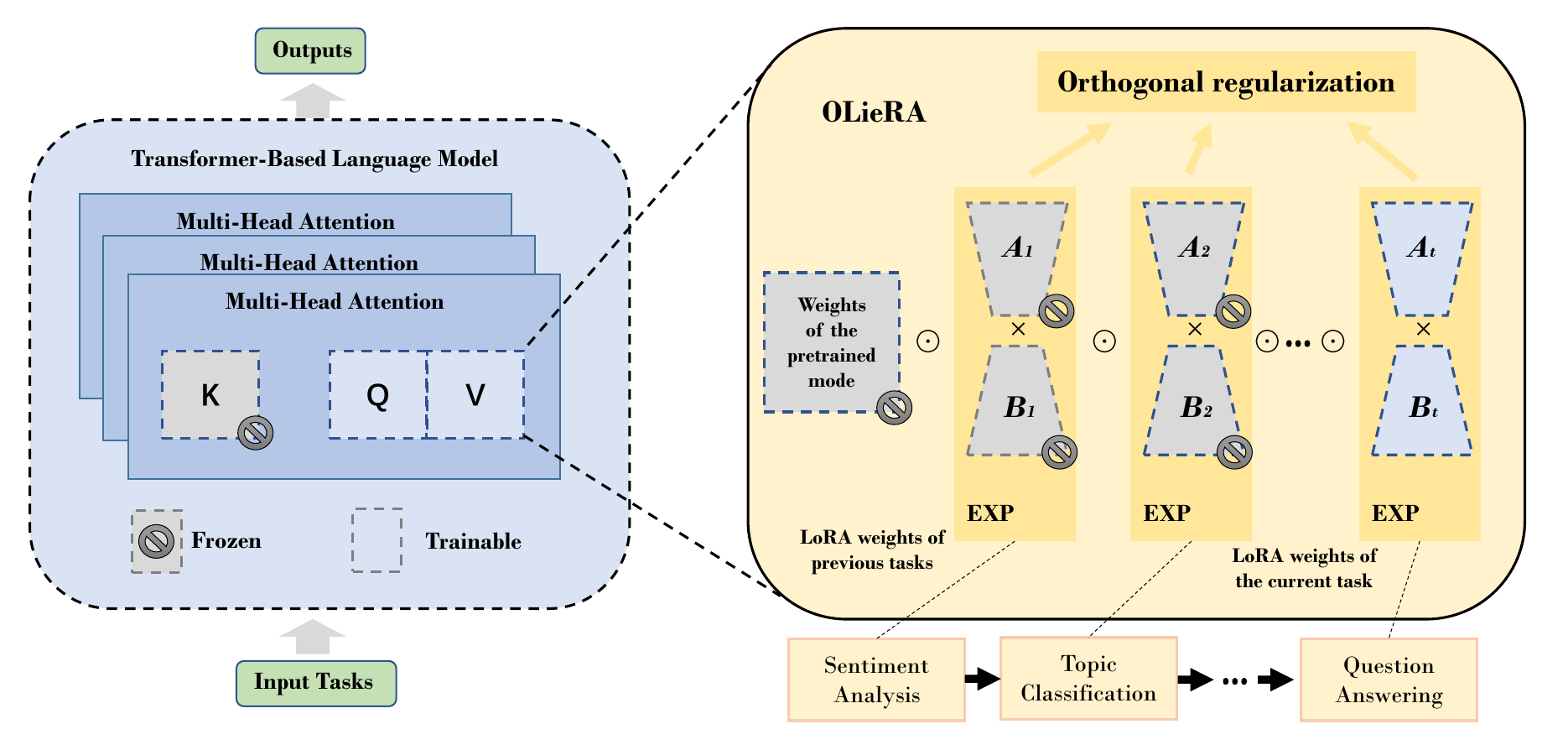}
    \caption{The OLieRA framework for continual learning in language models. First, human expertise is integrated and generalization is enhanced through instruction tuning. Second, building upon a frozen Transformer-based pre-trained language model, we leverage the exponential map of Lie groups to achieve "smooth manifold" parameter updates while preserving the intrinsic structure of the model. Then, we approximate the gradient subspace of each task using LoRA to efficiently control parameter overhead. For each sequentially arriving task, we incrementally learn a new LoRA module under both the Lie group constraint and an orthogonality constraint between the current task's LoRA subspace and all previous task subspaces, thereby reducing inter-task interference and preserving generalization to unseen tasks.}
    \label{fig:structure}
\end{figure*}

In this section, we introduce the theoretical framework of OLieRA, whose structure is illustrated in Figure~\ref{fig:structure}. First, we adopt instruction tuning as the training paradigm. Second, we learn new tasks via incremental exponential maps under Hadamard product in orthogonal subspaces, while keeping the LoRA parameters of completed tasks fixed. In the next section, we provide a comparative analysis between the proposed method and existing approaches.

\subsection{Introducing Lie Groups}

Lie groups have been widely applied in machine learning for manifold optimization, enabling parameter updates that respect intrinsic geometric structure \cite{leake2021optimization}. Representative applications include orthogonal parameterization \cite{plumbley2004lie} and modeling geometric transformations in computer vision \cite{liu2023lie,shutty2023computing}. In this work, we extend Lie group methods to the continual learning setting.  To model parameter updates via Lie theory, we embed the parameter tensor space into a Lie group. Let
\begin{equation}
\begin{split}
G = &\{ W \in \mathbb{R}^{b_1 \times \cdots \times b_k} \mid W[a_1,\ldots,a_k] \neq 0 \\
&\forall (a_1,\ldots,a_k) \},
\end{split}
\end{equation}
and define the group operation as the Hadamard product $\odot$. Requiring all entries to be non-zero ensures well-defined inverses and simplifies the structure \cite{varadarajan2013lie}.

Under $\odot$, $G$ is closed, associative, and commutative; the all-ones tensor serves as the identity, and each element admits an element-wise inverse. Since the operation is smooth, $(G, \odot)$ is an Abelian Lie group, isomorphic to a product of copies of $(\mathbb{R}^\ast, \times)$.

\subsection{Introducing Lie Algebras}

Viewing the Lie group $(G,\odot)$ as a smooth manifold, a parameter set $W \in G$ corresponds to a point on this manifold, while the associated Lie algebra $\mathfrak{g}$ is the tangent space at $W$ (Fig.~\ref{fig:Update Methods}). The Lie algebra is a linear space, where updates are simpler due to the availability of addition and scalar multiplication.

The Lie algebra is introduced because direct optimization on the Lie group is inconvenient. In principle, group updates take the form $W_{\text{new}} = W_{\text{old}} \odot \Delta W$ with all variables in $G$. However, gradient-based optimization typically computes additive updates, requiring the inversion
\begin{equation}
\Delta W = W_{\text{old}}^{-1} \odot W_{\text{new}},
\end{equation}
which complicates learning on the group. Instead, we perform updates in the Lie algebra $\mathfrak{g}$ and map them back to the Lie group via a suitable mapping (e.g., the exponential map), thereby recovering standard tensor operations compatible with frameworks such as PyTorch.

\subsection{Improved Incremental Update Mechanism}

Recent studies~\cite{marchetti2025position,modell2025originsrepresentationmanifoldslarge} suggest that neural network parameters lie on a largely smooth manifold, which can be interpreted as a Lie group. Given $W \in G$, we introduce an infinitesimal perturbation $\Delta W \in \mathfrak{g}$, perform linear updates in the Lie algebra, and map the result back to the group via the exponential map $\exp:\mathfrak{g}\!\to\!G$. Direct group-wise updates (e.g., $W \odot \Delta W$) are ill-posed, as $\Delta W$ may violate the constraints of $G$. The exponential map thus provides a principled bridge from $\mathfrak{g}$ to $G$.

Specifically, $\exp(\Delta W)$ (i) ensures local validity by mapping infinitesimal perturbations to admissible group elements, (ii) preserves structure by remaining within $G$ through element-wise exponential scaling, and (iii) maintains geometric consistency by aligning updates with the manifold geometry rather than arbitrary linear directions. Consequently, we adopt the update
\begin{equation}
W \rightarrow W \odot \exp(\Delta W),
\end{equation}

which combines structure-preserving group multiplication with efficient optimization in the linear Lie algebra, balancing geometric constraints and computational efficiency (Fig.~\ref{fig:Update Methods}).

\subsection{Taylor Approximation of the Exponential Map}

Since $\Delta W$ is small, the exponential map can be approximated using its first-order Taylor expansion:
\begin{equation}
\exp(\Delta W) = I + \Delta W + o(\|\Delta W\|)
\end{equation}
where $I$ denotes the matrix of all ones, and \(o(\|\Delta W\|)\) represents a remainder term. As \(\|\Delta W\|\) approaches zero, this remainder becomes negligible compared to \(\|\Delta W\|\) in an asymptotic sense. Therefore, we approximate the exponential map as \(\exp (\Delta W) \approx I + \Delta W\). Substituting this approximation into the multiplicative update yields:
\begin{equation}
W \odot \exp (\Delta W) \approx W \odot (I + \Delta W) = W + W \odot \Delta W
\end{equation}

For greater accuracy, we can employ a higher-order Taylor expansion of the exponential map:
\begin{equation}
\begin{split}
    \exp (\Delta W) = &I + \Delta W + \frac{1}{2} \Delta W \odot \Delta W +\\ &o(\| \Delta W \odot \Delta W \|)
\end{split}
\end{equation}
This is the second-order expansion. Essentially, this involves performing a higher-order Taylor expansion on each element individually.

\subsection{Low-Rank Decomposition}

We apply low-rank decomposition to $\Delta W$:
\begin{equation}
\Delta W = BA
\end{equation}
where $B \in \mathbb{R}^{\text{out} \times r}$, $A \in \mathbb{R}^{r \times \text{in}}$, and $r \ll \min(\text{out}, \text{in})$.

Substituting into the multiplicative update, we get:
\begin{equation}
\begin{split}
    W \odot \exp (\Delta W) &\approx W \odot (I + \Delta W)\\  &= W + W \odot (BA)
\end{split}
\end{equation}
Note: Matrix multiplication has higher precedence than the Hadamard product.

\subsection{Orthogonality Loss Calculation}

To mitigate catastrophic forgetting, we constrain the parameter space of each update to be as orthogonal as possible to the parameter spaces of all previous updates. Importantly, unlike \cite{wang-etal-2023-orthogonal}, our method applies the orthogonality constraint not just to the matrices $B$, but to the entire final updated parameter space, i.e., to $\exp(\Delta W)$.

Thus, we add a new loss term as follows:
\begin{equation}
\begin{split}
    \mathcal{L}_\text{orth} &= \sum_{i \neq j} \left\| \exp(\Delta W_i) \exp(\Delta W_j)^\top \right\|_F\\ &\approx \sum_{i \neq j} \left\| (I + B_iA_i) (I + B_jA_j)^\top \right\|_F
\end{split}
\end{equation}
where:
\begin{itemize}
    \item $\| \cdot \|_F$ denotes the Frobenius norm (i.e., $\sqrt{\sum_{k,l} |M_{kl}|^2}$)
    \item The summation is over all pairs of previous tasks for the model.
\end{itemize}

\subsection{Complete Formulation}

After incorporating the orthogonality constraint, the final total loss function is:
\begin{equation}
\mathcal{L}_\text{total} = \mathcal{L}_\text{task} + \lambda \sum_{i \neq j} \left\| (I + B_iA_i) (I + B_jA_j)^\top \right\|_F
\end{equation}
where $\lambda$ controls the strength of the orthogonality penalty term.

\section{Experimental Results}

\subsection{Experimental Setup}

\textbf{Datasets.}

To evaluate the effectiveness of our method, we follow O-LoRA \cite{wang-etal-2023-orthogonal} and N-LoRA \cite{yang-etal-2025-parameter} and utilize two benchmark datasets: 1) the Standard CL Benchmark, which comprises 5 text classification datasets, and 2) the Multi-Task Benchmark, which consists of 15 datasets, including those from GLUE, SuperGLUE, and IMDB. We adhere to the experimental protocols established by LFPT5 \cite{DBLP:journals/corr/abs-2110-07298}, O-LoRA \cite{wang-etal-2023-orthogonal}, and N-LoRA \cite{yang-etal-2025-parameter}, employing three different task sequences for each benchmark dataset. Further details are provided in the Appendix.

\textbf{Metrics.}

Let \( a_{i,j} \) denote the test accuracy on the \( i \)-th task after training on the \( j \)-th task. The evaluation metric adopted in this study is:

\textbf{Average Accuracy (AA)}, defined as the mean accuracy across all tasks after training on the final task is completed. It is calculated as \( A_T = \frac{1}{T} \sum_{i=1}^{T} a_{i,T} \), where \( T \) is the total number of tasks.

\textbf{Baseline Methods.}

To validate the effectiveness of OLieRA, we compare, as in O-LoRA \cite{wang-etal-2023-orthogonal} and N-LoRA \cite{yang-etal-2025-parameter}, against the following baseline methods:

\begin{itemize}
    \item[i.] \textbf{Isolated Training Methods}
    \begin{itemize}
        \item \textit{PerTaskFT}: Trains an independent model for each task.
        \item \textit{Prog-Prompt} \cite{razdaibiedina2023progressive}: Learns isolated prompts for each task, essentially equivalent to training a distinct model per task.
    \end{itemize}

    \item[ii.] \textbf{Upper Bound Method}
    \begin{itemize}
        \item \textit{Multi-Task Learning (MTL)}: Trains a single model on all tasks simultaneously. This method serves as the performance upper bound for continual learning.
    \end{itemize}

    \item[iii.] \textbf{Non-Continual Learning Methods}
    \begin{itemize}
        \item \textit{SeqFT} \cite{DBLP:journals/corr/abs-1906-01076}: Finetunes all model parameters sequentially across the task sequence.
        \item \textit{SeqLoRA}: Trains a single LoRA adapter across the task sequence while keeping the pre-trained model parameters frozen.
        \item \textit{IncLoRA}: Trains a new LoRA module for each task.
    \end{itemize}

    \item[iv.] \textbf{Continual Learning Methods}
    \begin{itemize}
        \item \textit{Replay}: Replays samples from previous tasks via a memory buffer while finetuning all model parameters.
        \item \textit{EWC} \cite{kirkpatrick2017overcoming}: Applies a Fisher regularization loss to adjust all model parameters.
        \item \textit{LwF} \cite{8107520}: Preserves knowledge from previous tasks by recording the outputs of the previous task's model on new task data and using these outputs as a regularization term.
        \item \textit{L2P} \cite{DBLP:journals/corr/abs-2112-08654} and \textit{LFPT5} \cite{DBLP:journals/corr/abs-2110-07298}: Both employ prompt-based mechanisms, adapting to new tasks by dynamically selecting or generating prompts.
        \item \textit{O-LoRA} \cite{wang-etal-2023-orthogonal}: Incrementally updates parameters for new tasks by constraining them to lie in the orthogonal subspace of parameters from previous tasks, while keeping the LoRA parameters learned from previous tasks fixed.
        \item \textit{N-LoRA} \cite{yang-etal-2025-parameter}: Learns parameters for new tasks in a non-parametric conflicting space under a stronger constraint than orthogonality (non-parametric conflict condition), while fixing the LoRA parameters learned previously, enabling incremental updates for new tasks.
    \end{itemize}
\end{itemize}

\begin{table*}[!h]
    \renewcommand{\arraystretch}{1}
    \setlength{\tabcolsep}{7pt}
    \renewcommand{\arraystretch}{0.7}
    \centering
    \begin{tabular}{c | c c c c | c c c c}
    \toprule
        Benchmarks & \multicolumn{4}{c|}{Standard CL Benchmark} & \multicolumn{4}{c}{Large Number of Tasks}  \\ 
        Methods & Order-1 & Order-2 & Order-3 & avg & Order-4 & Order-5 & Order-6 &  avg \\ 
        \toprule
        ProgPrompt & 75.2 & 75.0 & 75.1 & 75.1 & 78.0 & 77.7 & 77.9 & 77.9 \\ 
        PerTaskFT & 70.0 & 70.0 & 70.0 & 70.0 & 78.1 & 78.1 & 78.1 & 78.1 \\
        MTL & 80.0 & 80.0 & 80.0 & 80.0 & 76.5 & 76.5 & 76.5 & 76.5 \\
        \toprule
        SeqFT & 18.9 & 24.9 & 41.7 & 28.5 & 7.4 & 7.4 & 7.5 & 7.4 \\
        SeqLoRA & 44.6 & 32.7 & 53.7 & 43.7 & 0.6 & 1.9 & 1.6 & 1.6 \\
        IncLoRA & 66.0 & 64.9 & 68.3 & 66.4 & 63.3 & 58.5 & 61.7 & 61.2 \\
        Replay & 55.2 & 56.9 & 61.3 & 57.8 & 55.0 & 54.6 & 53.1 & 54.2 \\
        EWC & 48.7 & 47.7 & 54.5 & 50.3 & 45.3 & 44.5 & 45.6 & 45.1 \\
        LwF & 54.4 & 53.1 & 49.6 & 52.3 & 50.1 & 43.1 & 47.4 & 46.9 \\
        L2P & 60.3 & 61.7 & 61.1 & 60.7 & 57.5 & 53.8 & 56.9 & 56.1 \\
        LFPT5 & 67.6 & 72.6 & 77.9 & 72.7 & 70.4 & 68.2 & 69.1 & 69.2 \\ 
        O-LoRA & 75.4 & 75.7 & 76.3 & 75.8 & 72.3 & 64.8 & 71.6 & 69.6 \\
        N-LoRA & 79.2 & 78.4 & 78.8 & 78.8 & 73.6 & 70.3 & 73.2 & 72.4\\
        \textbf{OLieRA} & $\bm{79.9}$ & $\bm{79.5}$ & $\bm{79.5}$ & $\bm{79.6}$ & $\bm{73.8}$ & $\bm{70.4}$ & $\bm{73.5}$ & $\bm{72.6}$ \\
        \toprule
    \end{tabular}
    \caption{Summary of results on two standard continual learning benchmarks using the T5-large model. The average accuracy after training on the final task is reported, with all results averaged over three independent runs.}
    \label{table: 1}
\end{table*}

\begin{table}[!h]
    \renewcommand{\arraystretch}{1}
    \setlength{\tabcolsep}{2pt}
    \renewcommand{\arraystretch}{0.8}
    \centering
    \begin{tabular}{c | c c c c}
    \toprule
        Benchmarks & Order-1 & Order-2 & Order-3 & avg\\ 
        \toprule
        O-LoRA & 76.8 & 75.7 & 75.7 & 76.1\\ 
        N-LoRA & 77.2 & 77.3 & $\bm{78.4}$ & 77.6  \\ 
        OLieRA & $\bm{77.4}$ & $\bm{77.5}$ & 78.3 & $\bm{77.7}$  \\ 
        \toprule
    \end{tabular}
    \caption{Performance comparison between N-LoRA, O-LoRA and OLieRA on the larger LLaMA-7B model, reporting average accuracy across all task orders.  Results are averaged over three independent runs.}
    \label{table: 2}
\end{table}

\subsection{Comparison with State-of-the-Art Methods}

Tables~\ref{table: 1} and~\ref{table: 2} present the performance comparison, measured by Average Accuracy ($A_T$), between OLieRA and other baseline methods across different benchmark datasets. Following the experimental setup of LFPT5 \cite{DBLP:journals/corr/abs-2110-07298}, O-LoRA \cite{wang-etal-2023-orthogonal}, and N-LoRA \cite{yang-etal-2025-parameter}, we report the results from three independent runs with different task sequences and models on the Continual Learning (CL) benchmark dataset.

\textbf{Performance on the Standard Continual Learning Benchmark.}

OLieRA demonstrates stable and outstanding performance across the three task sequences (Order-1, Order-2, Order-3) of this benchmark: it achieves scores of 79.9, 79.5, and 79.5, respectively, with an average performance (avg) of 79.6.
This notably:
\begin{itemize}
    \item Surpasses the previous state-of-the-art (SOTA) method N-LoRA by 0.8\% (compared to N-LoRA's avg of 78.8).
    \item Outperforms isolated training methods such as ProgPrompt (avg 75.1) and PerTaskFT (avg 70.0).
    \item Performs exceedingly close to the continual learning upper bound—MTL (avg 80.0)—with a mere gap of 0.4\%, indicating that OLieRA can utilize knowledge from multiple tasks with high efficiency.
\end{itemize}
As shown in Table~\ref{table: 2}, OLieRA also significantly outperforms O-LoRA and mostly slightly surpasses N-LoRA on the larger-scale LLaMA-7B model. This validates that our proposed method remains effective even on larger-scale Large Language Models (LLMs).

\textbf{Performance in the Multi-Task Scenario.}

Continual learning involving a large number of tasks constitutes a significantly more challenging benchmark scenario. As shown in Table~\ref{table: 1}, under task sequences 4, 5, and 6, OLieRA's performance exceeds that of all existing state-of-the-art methods, achieving an average performance gain of 0.2\%. These results robustly demonstrate that OLieRA excels at preserving task-specific knowledge while handling a large number of tasks, exhibiting superior overall performance.

\section{Discussion}

\subsection{Interpretation via Lie Group Introduction}

Inspired by \cite{si-etal-2025-generalized}, we view the pre-trained parameters $W$ as elements of a Lie group, with fine-tuning updates $\Delta W$ interpreted as infinitesimal perturbations. Such perturbations naturally reside in the associated Lie algebra, i.e., the tangent space of the group. Through structure-preserving mappings such as the exponential map, these infinitesimal updates can be mapped back to the Lie group, ensuring that the updated parameters remain within the group. This framework addresses a fundamental issue: naive additive updates generally do not preserve Lie group structure. By constraining $\Delta W$ to satisfy Lie algebra conditions and applying the exponential map, we obtain principled, structure-preserving parameter updates.

This approach requires that $W$ encodes meaningful structure. Prior work has shown that convolutional kernels benefit from structure-preserving fine-tuning \cite{si-etal-2025-generalized}. We argue that Large Language Models (LLMs) similarly exhibit structures that should be preserved. Although these structures differ from the spatial locality of convolutional kernels and are harder to interpret, they are nonetheless critical. LLMs consist of heterogeneous yet highly structured components (e.g., Q/K/V matrices, feed-forward layers, and normalization parameters), whose internal organization must be maintained to avoid catastrophic forgetting. Moreover, these components are tightly coupled, and preserving their interdependencies is essential for stable adaptation.

Motivated by these observations, we adopt Lie group–based parameter updates of the form $W \odot \exp(\Delta W)$ instead of $W + \Delta W$. Importantly, $\Delta W$ can still be parameterized via low-rank decomposition inside $\exp(\Delta W)$, thereby retaining the efficiency and flexibility of low-rank adaptation while enforcing structural consistency.

\subsection{Order of Taylor Expansion}

To reduce computational overhead, we approximate the exponential map via a truncated Taylor expansion. Although this approximation is not strictly structure-preserving, higher-order expansions more closely match the ideal update, analogous to approximating a smooth curve with higher-degree polynomials. Hence, the expansion order directly controls the degree of structure preservation.

The $n$-th order Taylor expansion of the exponential map is
\begin{equation} 
\begin{split} 
\exp(\Delta W) &= I + \Delta W + \frac{1}{2!} \Delta W^{\odot 2} +\\ &\dots + \frac{1}{n!} \Delta W^{\odot n} + o(\|\Delta W^{\odot n}\|) 
\end{split} 
\end{equation}
where $I$ is the all-ones matrix and $\Delta W^{\odot 2}=\Delta W \odot \Delta W$ denotes the Hadamard product.

We validate this effect through an ablation study on the Standard CL Benchmark with T5-base, comparing first-, second-, and third-order approximations. Results averaged over three runs are reported in Table~\ref{table: 3}.

\begin{table}[!h]
    \renewcommand{\arraystretch}{1}
    \setlength{\tabcolsep}{2pt}
    \renewcommand{\arraystretch}{0.8}
    \centering
    \begin{tabular}{c | c c c c}
    \toprule
        Taylor order & 1st-order & 2nd-order & 3rd-order \\ 
        \toprule
        Order-1 & 79.4 & 79.9 & 79.9 \\ 
        Order-2 & 79.2 & 79.5 & 79.6  \\ 
        Order-3 & 79.3 & 79.5 & 79.4   \\ 
        \toprule
    \end{tabular}
    \caption{The performance of model on Standard CL Benchmark using the T5-large model with first-order, second-order and third-order Taylor approximation.  Results are averaged over three independent runs.}
    \label{table: 3}
\end{table}

\subsection{Effectiveness of the Multiplicative Update}
\label{subsec:abl_mult}

To assess the necessity of the Lie group–motivated multiplicative update, we perform an ablation study in which the proposed update $W \odot \exp(\Delta W)$ is replaced with a standard additive update $W + \Delta W$, while keeping the subspace orthogonality constraint unchanged. We refer to this variant as \textit{No LieGroup Mult}. Results are reported in Table~\ref{table: 8}.

Removing the multiplicative update leads to a consistent and substantial performance drop across all task sequences on the Standard CL Benchmark, with an average degradation of approximately $2.3\%$. This confirms that the exponential map–based multiplicative update is a core component of OLieRA rather than a replaceable implementation detail, and that it works synergistically with the orthogonality constraint.

We attribute this effect to the scaling nature of the multiplicative update, which better preserves the intrinsic structure of pre-trained weights and provides a more stable parameter space for subsequent orthogonally constrained adaptations. Without this structure-preserving mechanism, the benefit of the subspace orthogonality constraint is significantly reduced.

\begin{table}[!h]
    \renewcommand{\arraystretch}{1}
    \setlength{\tabcolsep}{2pt}
    \renewcommand{\arraystretch}{0.8}
    \centering
    \begin{tabular}{c | c c c c}
    \toprule
        Fisher value & Order-1 & Order-2 & Order-3 \\ 
        \toprule
        OLieRA & 79.9 & 79.5 & 79.5 \\ 
        No LieGroup Mult & 77.4 & 77.2 & 76.9   \\ 
        \toprule
    \end{tabular}
    \caption{The ablation experiment of effectiveness of multiplicative updates on Standard CL Benchmark using the T5-large model. Results are averaged over three independent runs.}
    \label{table: 8}
\end{table}

\subsection{Unveiling the Internal Mechanism via Fisher Information}
In continual learning, the Fisher Information Matrix (FIM) is widely used to measure the importance of parameters for previously learned tasks. For model parameters $\theta$, the $i$-th diagonal element is empirically approximated as
\begin{equation}
F_i \approx \mathbb{E}_{x \sim \mathcal{D}} \left[ \left( \partial_{\theta_i} \log p(y \mid x; \theta) \right)^2 \right],
\end{equation}
where larger $F_i$ indicates greater sensitivity of old-task performance to $\theta_i$, and thus stronger protection during fine-tuning.

In LoRA-based continual learning, each task induces a parameter update $\Delta \theta$. Under the diagonal FIM approximation \cite{kirkpatrick2017overcoming}, the conflict between new and old tasks can be quantified by the Fisher-weighted energy
\begin{equation}
E = \sum_i F_i (\Delta \theta_i)^2 ,
\end{equation}
which measures how strongly the new update perturbs directions important to previous tasks.

\begin{table}[!h]
    \renewcommand{\arraystretch}{1}
    \setlength{\tabcolsep}{2pt}
    \renewcommand{\arraystretch}{0.8}
    \centering
    \begin{tabular}{c | c c c c}
    \toprule
        Fisher value & Order-1 & Order-2 & Order-3 \\ 
        \toprule
        O-LoRA & 0.12 & 0.09 & 0.42 \\ 
        N-LoRA & 7e-11 & 8e-11 & 1e-10  \\ 
        OLieRA & 1.04 & 1.43 & 3.92   \\ 
        \toprule
    \end{tabular}
    \caption{The Fisher value of model on Standard CL Benchmark using the T5-large model with different methods.  Results are averaged over three independent runs.}
    \label{table: 7}
\end{table}

We compute this energy between the last two tasks (Table~\ref{table: 7}). Consistent with prior work \cite{yang-etal-2025-parameter}, N-LoRA yields near-zero energy, indicating strong orthogonality, while O-LoRA produces larger values. In contrast, OLieRA exhibits higher Fisher-weighted energy than both baselines, suggesting that its Lie group–based updates do not simply avoid sensitive directions but enable controlled updates along them. We argue that this improves parameter utilization and knowledge sharing, resulting in greater expressiveness and superior overall performance despite stronger interaction with high-Fisher directions.

\section{Conclusion}

Through both theoretical derivation and empirical validation, this work demonstrates that enforcing orthogonality constraints on the full task-update subspace—while preserving the inherent structure of pretrained model parameters—is a key strategy for mitigating catastrophic forgetting in large language models (LLMs). Building on this insight, we propose \textbf{OLieRA} (Orthogonal Low-rank adaptation in Lie groups), a method that combines theoretical interpretability with practical efficiency. By integrating a multiplicative update paradigm grounded in Lie group theory (which preserves the geometric structure of parameters), full-subspace orthogonality constraints (which reduce task interference), and Taylor approximation techniques (which balance accuracy and computational cost), OLieRA achieves a new breakthrough in continual learning performance.

Comprehensive experiments on both the standard CL benchmark (5 text classification tasks) and the long-sequence multi-task benchmark (15 tasks spanning GLUE and SuperGLUE) demonstrate that, compared with state-of-the-art methods such as O-LoRA and N-LoRA, OLieRA not only maintains parameter structure while enforcing full-subspace orthogonality, but also delivers superior average accuracy (79.6\% on the standard CL benchmark, approaching the MTL upper bound; 72.6\% on the multi-task benchmark, surpassing existing methods). Moreover, OLieRA provides stronger resistance to task interference while retaining the key advantages of O-LoRA—no need for historical data storage (privacy-friendly), only a small number of additional parameters (cost-efficient), and independence from task IDs at test time (compatible with instruction-tuning generalization). Finally, Fisher information analysis further confirms that OLieRA enables controlled updates along sensitive parameter directions of prior tasks, effectively balancing task conflict with knowledge sharing, and offering a novel paradigm toward practical continual learning for LLMs.

\section{Ethics and AI Disclosure}

We used ChatGPT to assist with text drafting. All scientific content and experiments were independently done by the authors. 
\bibliography{main}
  
\clearpage
\appendix

\section{Appendix}
\label{sec:appendix}

\subsection{Experimental Details}
All experiments with the T5 model were conducted on a single machine equipped with an NVIDIA A100-SXM4-80GB GPU, and implemented based on the DeepSpeed library. For all task order sequences, we trained the models with the following unified hyperparameters: the number of training epochs was set to 2, the learning rate was fixed at 1e-3 for the Standard CL Benchmark and 5e-4 for the Large Number of Tasks setting, the dropout rate was set to 0.1, and the weight decay rate was set to 0. The only exceptions occur in Orders 1–6, where the values of $\lambda_1$ and $\lambda_2$ differ as follows:

For Order-1, Order-2, and Order-3, we set $\lambda_1=0.5$ and $\lambda_2=0$.  
For Order-4 (with the task sequence MNLI, CB, WiC, COPA, QQP, BoolQA, RTE, IMDB, Yelp, Amazon, SST-2, DBpedia, Agnews, MultiRC, Yahoo), the values of $\lambda_1$ were set to 0.5, 0.5, 0.5, 0.5, 0.5, 0.5, 0.5, 0.5, 0.5, 0.5, 0.5, 0.6, 0.7, 0.4, 0.6, and $\lambda_2$ to 0, 0, 0.1, 0, 0, 0, 0, 0, 0, 0, 0, 0.01, 0.02, 0, 0.05.  
For Order-5 (with the task sequence MultiRC, BoolQA, WiC, MNLI, CB, COPA, QQP, RTE, IMDB, SST-2, DBpedia, Agnews, Yelp, Amazon, Yahoo), the values of $\lambda_1$ were set to 0.7, 0.7, 0.7, 0.7, 0.7, 0.7, 0.7, 0.7, 0.7, 0.7, 0.7, 0.5, 0.7, 0.4, 1, and $\lambda_2$ to 0, 0, 0, 0, 0, 0, 0, 0, 0, 0, 0, 0.05, 0.04, 0.08, 0.15.  
For Order-6 (with the task sequence Yelp, Amazon, MNLI, CB, COPA, QQP, RTE, IMDB, SST-2, DBpedia, Agnews, Yahoo, MultiRC, BoolQA, WiC), $\lambda_1$ was uniformly set to 1.1 and $\lambda_2$ to 0.  

For experiments with the LLAMA-7B model, we adopted the following settings:  
For all orders, $\lambda_1$ was set to 0.5 and $\lambda_2$ to 0.  
For tasks in Order-1, the learning rates were set to (1e-3, 1e-4, 1e-4, 1e-4), batch sizes to (8, 4, 8, 8), dropout rate to 0.1, weight decay to 0, and the number of training epochs to (1, 2, 1, 2).  
For tasks in Order-2, the learning rates were set to (1e-3, 1e-4, 1e-4, 1e-4), batch sizes to (8, 4, 8, 8), dropout rate to 0.1, weight decay to 0, and the number of training epochs to (1, 2, 2, 1).  
For tasks in Order-3, the learning rates were set to (1e-3, 1e-4, 1e-4, 1e-4), batch sizes to (8, 4, 8, 8), dropout rate to 0.1, weight decay to 0, and the number of training epochs to (1, 2, 2, 1).  

\subsection{Datasets}
Table~\ref{table: 4} provides detailed descriptions of the 15 datasets and their evaluation metrics used in our continual learning (CL) experiments. Overall, the datasets were drawn from three main benchmark suites: the CL benchmark~\cite{zhang2015character}, GLUE~\cite{wang2018glue}, and SuperGLUE~\cite{wang2019superglue}. In addition, following~\cite{razdaibiedina2023progressive}, we supplemented these benchmarks with the IMDB movie reviews dataset. Data is secure and has been anonymized.

\begin{table*}[t]
\begin{tabular}{l|llll}
\hline
\textbf{Dataset name} & \textbf{Category} & \textbf{Task}             & \textbf{Domain}     & \textbf{Metric} \\ \hline
1. Yelp               & CL Benchmark      & sentiment analysis        & Yelp reviews        & accuracy        \\
2. Amazon             & CL Benchmark      & sentiment analysis        & Amazon reviews      & accuracy        \\
3. DBpedia            & CL Benchmark      & topic classification      & Wikipedia           & accuracy        \\
4. Yahoo              & CL Benchmark      & topic classification      & Yahoo Q\&A          & accuracy        \\
5. AG News            & CL Benchmark      & topic classification      & news                & accuracy        \\
6. MNLI               & GLUE              & NLI                       & various             & accuracy        \\
7. QQP                & GLUE              & paragraph detection       & Quora               & accuracy        \\
8. RTE                & GLUE              & NLI                       & news, Wikipedia     & accuracy        \\
9. SST-2              & GLUE              & sentiment analysis        & movie reviews       & accuracy        \\
10. WiC               & SuperGLUE         & word sense disambiguation & lexical databases   & accuracy        \\
11. CB                & SuperGLUE         & NLI                       & various             & accuracy        \\
12. COPA              & SuperGLUE         & QA                        & blogs, encyclopedia & accuracy        \\
13. BoolQA            & SuperGLUE         & boolean QA                & Wikipedia           & accuracy        \\
14. MultiRC           & SuperGLUE         & QA                        & various             & accuracy        \\
15. IMDB              & SuperGLUE         & sentiment analysis        & movie reviews       & accuracy        \\ \hline
\end{tabular}
\caption{The details of 15 datasets used in our CL experiments. NLI denotes natural language
inference, QA denotes questions and answers task. First five tasks
correspond to the standard CL benchmark, all other tasks are used in long-sequence experiments.
}
\label{table: 4}
\end{table*}

\subsection{Task Sequence Orders}
We report the task sequences used in our CL experiments in Table~\ref{table: 5}.

\begin{table*}[h]
\begin{tabular}{lll}
\hline
\textbf{Order} & \textbf{Model} & \textbf{Task Sequence}                                                                                                                                \\ \hline
1              & T5, LLaMA      & dbpedia $ \to $ amazon $ \to $ yahoo $ \to $ ag                                                                                                                         \\
2              & T5, LLaMA      & dbpedia $ \to $ amazon $ \to $ ag $ \to $ yahoo                                                                                                                         \\
3              & T5, LLaMA      & yahoo $ \to $ amazon $ \to $ ag $ \to $ dbpedia                                                                                                                         \\ \hline
4              & T5             & \begin{tabular}[c]{@{}l@{}}mnli $ \to $ cb $ \to $ wic $ \to $ copa $ \to $ qqp $ \to $ boolqa $ \to $ rte $ \to $ imdb $ \to $\\ yelp $ \to $ amazon $ \to $ sst-2 $ \to $ dbpedia $ \to $ ag $ \to $ multirc $ \to $ yahoo\end{tabular} \\
5              & T5             & \begin{tabular}[c]{@{}l@{}}multirc $ \to $ boolqa $ \to $ wic $ \to $ mnli $ \to $ cb $ \to $ copa $ \to $ qqp $ \to $ rte\\ $ \to $ imdb $ \to $ sst-2 $ \to $ dbpedia $ \to $ ag $ \to $ yelp $ \to $ amazon $ \to $ yahoo\end{tabular} \\
6              & T5             & \begin{tabular}[c]{@{}l@{}}yelp $ \to $ amazon $ \to $ mnli $ \to $ cb $ \to $ copa $ \to $ qqp $ \to $ rte $ \to $ imdb $ \to $\\ sst-2 $ \to $ dbpedia $ \to $ ag $ \to $ yahoo $ \to $ multirc $ \to $ boolqa $ \to $ wic\end{tabular} \\ \hline
\end{tabular}
\caption{Six different orders of task sequences used for continual learning experiments. Orders
1-3 correspond to the standard CL becnhmark adopted by prior works. Orders 4-6 are long-sequence orders spanning 15 tasks, following \cite{razdaibiedina2023progressive}.}
\label{table: 5}
\end{table*}

\subsection{Task Instructions}
Table~\ref{table: 6} presents the prompts used for different tasks.  
NLI denotes Natural Language Inference, including MNLI, RTE, and CB.  
SC denotes Sentiment Classification, including Amazon, Yelp, SST-2, and IMDB.  
TC denotes Topic Classification, including AGNews, Dbpedia, and Yahoo.

\begin{table*}[h]
\begin{tabular}{cl}
\hline
\textbf{Task}                                                       & \multicolumn{1}{c}{\textbf{Prompts}}                                                                                                                                  \\ \hline
NLI                                                                 & \begin{tabular}[c]{@{}l@{}}What is the logical relationship between the "sentence 1" and the "sentence 2"? \\ Choose one from the option.\end{tabular}                \\ \hline
QQP                                                                 & \begin{tabular}[c]{@{}l@{}}Whether the "first sentence" and the "second sentence" have the same meaning? \\ Choose one from the option.\end{tabular}                  \\ \hline
\begin{tabular}[c]{@{}c@{}}SC\end{tabular}   & What is the sentiment of the following paragraph? Choose one from the option.                                                                                         \\ \hline
\begin{tabular}[c]{@{}c@{}}TC\end{tabular} & What is the topic of the following paragraph? Choose one from the option.                                                                                             \\ \hline
BoolQA                                                              & \begin{tabular}[c]{@{}l@{}}According to the following passage, is the question true or false? Choose one \\ from the option.\end{tabular}                             \\ \hline
MultiRC                                                             & \begin{tabular}[c]{@{}l@{}}According to the following passage and question, is the candidate answer true \\ or false? Choose one from the option.\end{tabular}        \\ \hline
WiC                                                                 & \begin{tabular}[c]{@{}l@{}}Given a word and two sentences, whether the word is used with the same sense \\ in both sentence? Choose one from the option.\end{tabular} \\ \hline
\end{tabular}
\caption{Instructions for different tasks.}
\label{table: 6}
\end{table*}
\subsection{Limitations}

Despite its empirical effectiveness, OLieRA has several limitations that warrant further investigation. 

First, the proposed formulation models model parameters as elements of an Abelian Lie group defined by the Hadamard product. While this choice enables efficient and stable multiplicative updates, it remains unclear to what extent faithfully reflects structures that may underlie large language model parameters.

Second, compared with strong baselines such as O-LoRA and N-LoRA, the performance improvement brought by our proposed method is relatively modest in long task sequences, which may imply that more sophisticated approaches are required to maintain the parameter structure for long task sequences.

Finally, the method introduces additional hyperparameters (e.g., the strength of orthogonality regularization), whose tuning may be nontrivial and task-order dependent.

Addressing these limitations, particularly through richer geometric parameterizations and more principled approximation schemes, is an important direction for future work.

Potential risks of our continual learning approach include performance bias across tasks or populations, privacy concerns when models are trained on sensitive data, and possible misuse in high-stakes decision settings if the model is applied outside controlled experiments.

\end{document}